\documentclass[journal]{IEEEtran}

\usepackage{float}
\usepackage{graphicx}
\usepackage{multirow}
\usepackage{amsmath} 
\usepackage{textcomp}
\usepackage{tabularx}
\usepackage[normalem]{ulem}
\usepackage{booktabs}
\usepackage{footnote}
\usepackage{threeparttable}
\usepackage[colorlinks,linkcolor=red]{hyperref}
\usepackage{cite}
\usepackage{booktabs}
\usepackage{caption}
\usepackage[utf8]{inputenc}
\usepackage{amssymb}             
\usepackage{amsfonts}            
\usepackage{mathrsfs} 
\usepackage{CJKutf8}
\usepackage[table]{xcolor}
\usepackage[noend]{algpseudocode}
\usepackage{booktabs}
\usepackage{indentfirst}
\usepackage{threeparttable}
\usepackage[ruled,linesnumbered]{algorithm2e}
\usepackage{array}
\usepackage{stfloats}
\usepackage{url}
\usepackage{verbatim}
\usepackage{graphicx}
\usepackage{float}
\usepackage{booktabs} 
\usepackage{subfigure}
\usepackage{cite}
\usepackage{colortbl}
\usepackage{cleveref}
\usepackage{makecell}
\usepackage{rotating}
\usepackage{subcaption}

\usepackage[colorinlistoftodos]{todonotes}

\begin{document}

\title{ContrastAlign: Toward Robust BEV Feature Alignment via Contrastive Learning for Multi-Modal 3D Object Detection}

\author{Ziying Song, Hongyu Pan, Feiyang Jia, Yongchang Zhang, Lin Liu,\\  Lei Yang, Shaoqing Xu, Peiliang Wu, Caiyan Jia, Zheng Zhang,  Yadan Luo
\thanks{This work was supported by the National Key R\&D Program of China (2018AAA0100302).\emph{(Corresponding author: Caiyan Jia.)}}
\thanks{Ziying Song, Feiyang Jia, Lin Liu, Caiyan Jia are with School of Computer Science \& Technology, Beijing Key Laboratory of Traffic Data Mining and Embodied Intelligence, Beijing Jiaotong University (e-mail: 22110110@bjtu.edu.cn, feiyangjia@bjtu.edu.cn, liulin010811@gmail.com,  cyjia@bjtu.edu.cn)
}
\thanks{Hongyu Pan and Yongchang Zhang are with Horizon Robotics (e-mail:
karry.pan@horizon.auto, zhangyongchang20@mails.ucas.ac.cn).
}
\thanks{Lei~Yang is with the  Nanyang Technological University, Singapore. (email: yangleils@outlook.com).}

\thanks{Shaoqing Xu is with  University
of Macau, China (e-mail: shaoqing.xu@connect.um.edu.mo)
}
\thanks{Peiliang Wu is with School of Information Science and Engineering, Yanshan University, Qinhuangdao 066004, China (e-mail: peiliangwu@ysu.edu.cn;)}
\thanks{Zheng Zhang is with the School of Computer Science and Technology, Harbin Institute of Technology, Shenzhen, Guangdong 518055, China (e-mail: darrenzz219@gmail.com).}
\thanks{Yadan Luo is with the School of Information Technology and Electrical
Engineering, The University of Queensland, Australia
(e-mail: uqyluo@uq.edu.au)
}

}

\maketitle

\begin{abstract}
In the field of 3D object detection tasks, fusing heterogeneous features from LiDAR and camera sensors into a unified Bird's Eye View (BEV) representation is a widely adopted paradigm. However, existing methods often suffer from imprecise sensor calibration, leading to feature misalignment in LiDAR-camera BEV fusion. Moreover, such inaccuracies cause errors in depth estimation for the camera branch, aggravating misalignment between LiDAR and camera BEV features. In this work, we propose a novel \textbf{ContrastAlign} approach that utilizes contrastive learning to enhance the alignment of heterogeneous modalities, thereby improving the robustness of the fusion process. Specifically, our approach comprises three key components: (1) the L-Instance module, which extracts LiDAR instance features within the LiDAR BEV features; (2) the C-Instance module, which predicts camera instance features through Region of Interest (RoI) pooling on the camera BEV features; (3) the InstanceFusion module, which employs contrastive learning to generate consistent instance features across heterogeneous modalities. Subsequently, we use graph matching to calculate the similarity between the neighboring camera instance features and the similarity instance features to complete the alignment of instance features. Our method achieves SOTA performance, with an mAP of 71.5\%, surpassing GraphBEV by 1.4\%  on the nuScenes val set. Importantly, our method excels BEVFusion under conditions with spatial \& temporal misalignment noise, improving mAP by 1.4\% and 11.1\% on nuScenes dataset. Notably, on the Argoverse2 dataset, ContrastAlign outperforms GraphBEV by 1.0\% in mAP, indicating that the farther the distance, the more severe the feature misalignment and the more effective.

\end{abstract}


\IEEEpeerreviewmaketitle

\section{Introduction}

\begin{figure}[!t]
\includegraphics[width=1\linewidth]{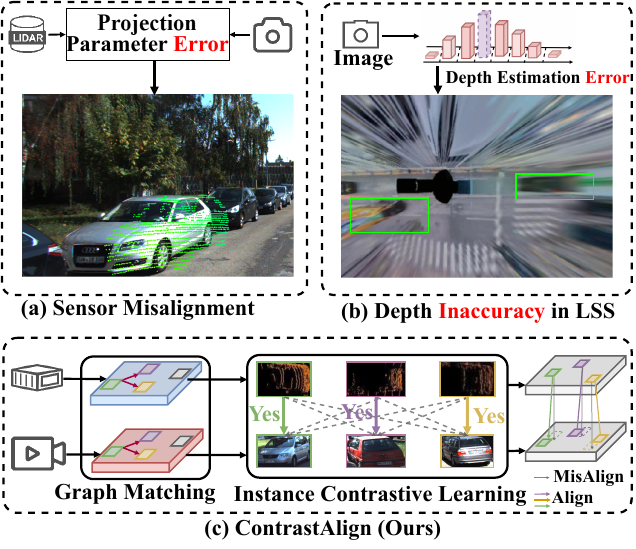}
\caption[ ]{Motivation of \textbf{ContrastAlign}. 
(a) Misalignment persists in real-world scenarios, particularly during vibrations while driving on bumpy roads. Since such noise occurs frequently, it cannot be easily mitigated through online calibration. (b) During the fusion of LiDAR BEV and camera BEV in BEVFusion ~\cite{bevfusion-mit,bevfusion-pku}, the errors in depth within Camera-to-BEV modules (e.g., LSS  ~\cite{lss}) lead to misalignment of BEV features. 
(c) 
Our contrastAlign framework achieves BEV feature alignment by leveraging contrastive learning to calculate the similarity between heterogeneous instance features of LiDAR and camera. In the BEV fusion stage, the direct concatenation method of BEVFusion may have misalignment issues (gray solid line), but our ContrastAlign corrects this problem through instance alignment.

} 
\label{fig:motivation}
\end{figure}

\IEEEPARstart{3}{D} object detection is one of the fundamental tasks for achieving reliable environment perception in autonomous driving tasks \cite{song2024robustness,zhao2023comprehensive,feng2021point_TCybernetics,survey_TCybernetics}. The task aims to accurately identify and locate obstacles such as cars and pedestrians, to provide precise and real-time data that informs the autonomous driving system 
to reach correct driving decisions \cite{zhou2025state_TCybernetics}. The current standard for achieving safe, robust, and high-precision detection may require the fusion of heterogeneous modalities. Due to the inherent sparsity of point clouds, LiDAR-only methods 
\cite{Pointpillars,Second,Voxelnet,Voxelrcnn,Pointrcnn,pvcnn,SATGCN,vpnet,xu2024tigdistill,chen2024lvp,yang2024t3dnet_TCybernetics,liu2022fg_TCybernetics,wang2020real_TCybernetics}
struggle to detect small or distant objects, making them insufficient for robust 3D detection. In contrast, these objects remain clear and distinguishable in high-resolution images, containing rich semantic representations \cite{wang2023multi}. The complementary roles of point clouds and images have prompted researchers to leverage the strengths of heterogeneous modalities to design detectors.

Based on different fusion strategies, heterogeneous 3D object detection can be mainly categorized into point-level  \cite{mvx-net, pointpainting, epnet, epnet++, wang2021pointaugmenting}, feature-level \cite{Transfusion, DeepFusion, DeepInteraction, autoalignv2, graphalign,xu2024sparseinteraction, graphalign++,song2024robofusion,VoxelNextFusion} methods and the currently dominant BEV-based methods  \cite{bevfusion-mit, bevfusion-pku}. BEV-based methods integrate LiDAR and camera modalities into a shared BEV representation space.
Although BEV-based methods \cite{bevfusion-mit, bevfusion-pku} have demonstrated promising performance, 
they still suffer \textbf{BEV feature misalignment}. There are two main causes of this problem, including sensor misalignment (see Figure \ref{fig:motivation} (a)) and depth inaccuracy (see Figure \ref{fig:motivation} (b)). Firstly, for sensor misalignment, as pointed out in BEVDepth \cite{bevdepth}, GraphBEV \cite{song2024graphbev}, ObjectFusion \cite{ObjectFusion}, calibration matrix errors between LiDAR and camera sensors can result in feature misalignment. Second, their fusion process relies heavily on the accuracy of depth estimation from camera to BEV (e.g., LSS\cite{lss}), with depth inaccuracy leading to  feature misalignment further.

The key to achieving feature alignment in autonomous driving lies in the deviation of the projection matrix (calibration matrix), which poses a challenge in the real world. Some feature-level methods  \cite{autoalign, Transfusion, DeepFusion, DeepInteraction, bi2024dyfusion} achieve feature fusion by cross-attention querying image features with point cloud features without the need for the projection matrix, but have significant computational overhead. Other feature-level methods  \cite{autoalignv2, graphalign, graphalign++, HMFI} attempt to alleviate alignment errors caused by feature alignment through Deformable Attention \cite{deformabledetr} and neighboring projections at the help of the projection matrix. As described by BEVFusion  \cite{bevfusion-mit}, although LiDAR BEV features and camera BEV features are in the same space, due to inaccuracies in the depth of the viewpoint transformer, they can still be spatially misaligned to some extent. 
So far, only a few BEV-based works  \cite{ObjectFusion,song2024graphbev} address the issue of BEV space feature misalignment. Where ObjectFusion \cite{ObjectFusion} proposes a novel object-centric fusion  to align object-centric features of different modalities, 
GraphBEV  \cite{song2024graphbev} mitigates misalignment issues by matching neighbor depth features through graph matching.

The issue of feature alignment exists not only in multi-modal 3D object detection  ~\cite{pointpainting,wang2021pointaugmenting,bevfusion-mit,bevfusion-pku} but also in multi-modal tasks involving text and images  ~\cite{zhan2023multimodal,xu2023multimodal}. With the development of multi-modal foundational models, more researchers  ~\cite{clip,wang2021actionclip,trex2,khattak2023maple,carlini2024aligned} are paying attention to heterogeneous modality alignment for the purpose of modality consistency. Contrastive alignment  ~\cite{trex2,zhan2023multimodal,xu2023multimodal} can be seen as a mutually refining process where each modality contributes to and benefits from the exchange of knowledge. Through contrastive learning, the iterative interaction between heterogeneous modalities continuously evolves, enhancing their ability for general understanding within a single model and addressing the alignment issues of heterogeneous modalities ~\cite{trex2}. Therefore, inspired by the above idea, we apply its concept to the task of 3D object detection to solve the feature misalignment between LiDAR and camera BEV features.

In this work, we propose \textbf{ContrastAlign}, which leverages contrastive learning to enhance the alignment of heterogeneous modalities, thereby improving the robustness of LiDAR-camera BEV feature fusion as shown in Figure \ref{fig:motivation} (c). 
Specifically, we propose the L-Instance module, which directly outputs LiDAR instance features within LiDAR BEV features. Then, we introduce the C-Instance module, which predicts camera instance features through RoI (Region of Interest) Pooling on camera BEV features. The LiDAR instance features are then projected onto image instance features, and contrastive learning is employed to generate similar instance features between LiDAR and camera. Subsequently, through graph matching, neighboring camera instance features are matched to calculate similarity and construct positive and negative samples. During inference, the aligned features with high similarity in neighboring camera instance features are selected as alignment features to achieve BEV feature alignment.
Extensive experiments have demonstrated the effectiveness of our ContrastAlign, with significant performance improvement on the 
nuScenes  ~\cite{nuscenes} dataset, especially at the misaligned noisy setting  ~\cite{zhujun_benchmarking}.
This is noteworthy, as we focus on object-level alignment and effectively address temporal misalignment using only single-frame multi-modal data.

\begin{figure*}[t]
\centering
\includegraphics[width=1\textwidth]{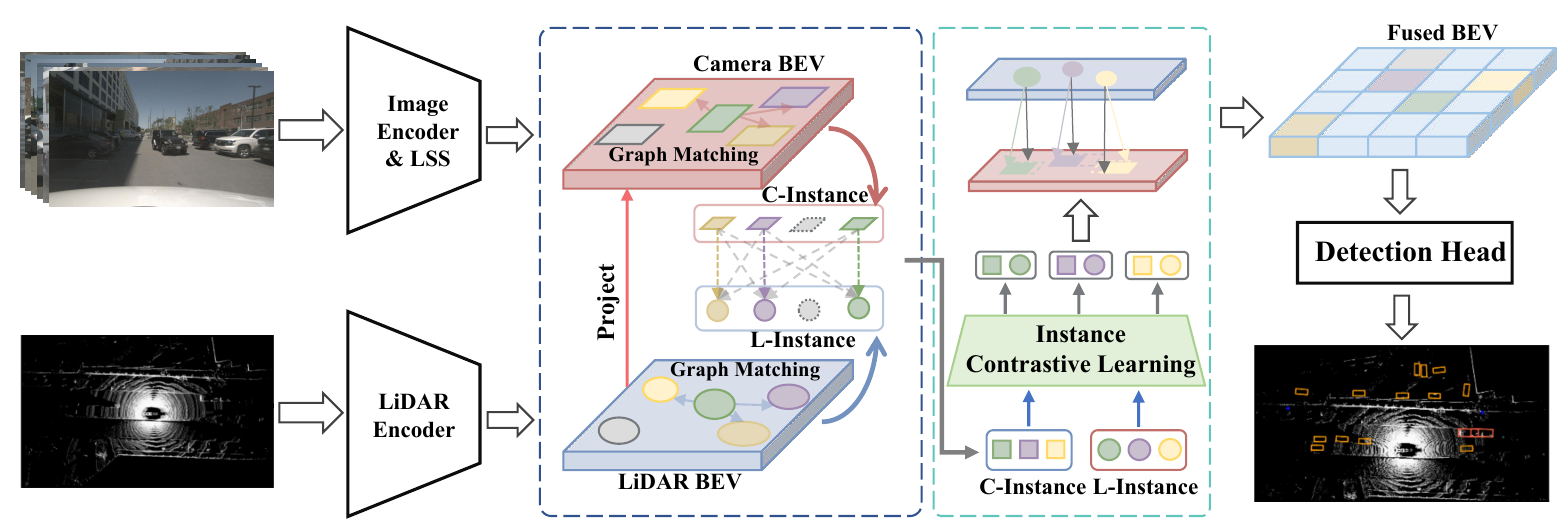}
\caption[ ]{Overview of our \textbf{ContrastAlign} framework. 
We follow the baseline~\cite{bevfusion-mit} to generate LiDAR BEV features in the LiDAR branch. Then, we propose the \textbf{L-Instance} module,  which directly outputs LiDAR instance features within LiDAR BEV features. Then, we introduce the \textbf{C-Instance} module, which predicts camera instance features through RoI (Region of Interest) Pooling on the camera BEV features. To align the Instance features of LiDAR and camera, we propose the \textbf{InstanceFusion} module. Sepiccfy, we establish neighbor relationships among instance features by employing graph matching techniques. Subsequently, we leverage contrastive learning to create pairs of positive and negative samples, facilitating learning similarities across diverse modalities. In the inference phase, we retrieve aligned image instance features by querying based on their shared characteristics. Finally, we employ a dense detection head  ~\cite{Transfusion} to accomplish the 3D detection task.
Notably, to preserve our core contributions better, we simplified the overview. For detailed modules such as LSS ~\cite{lss}, please refer to BEVFusion  ~\cite{bevfusion-mit}.

} 
\label{fig:framework}
\end{figure*}

\section{Related Work}

\subsection{Multi-modal 3D Object Detection}

Multi-modal 3D object detection has achieved state-of-the-art performance in KITTI, nuScenes, and other datasets by utilizing data features from heterogeneous sensors (LiDAR and camera) and integrating them to enhance the detection of 3D objects  ~\cite{song2024robustness, wang2023multi}. Multi-modal 3D object detectors can be broadly categorized into three fusion methods including point-level, feature-level and BEV-based methods. Point-level methods ~\cite{mvx-net,pointpainting,wang2021pointaugmenting,epnet,epnet++,mvp,xu2024multi} aim to enhance raw LiDAR points with image features and then pass them through a LiDAR-only 3D detector to produce 3D detection results. Feature-level methods  ~\cite{chen2022focalsconv,VoxelNextFusion,autoalign,autoalignv2,graphalign,graphalign++,Transfusion,DeepInteraction,bi2024dyfusion} primarily focus on integrating point cloud features with image features during the feature extraction stage. Among feature-level methods, representative works such as HMFI  ~\cite{HMFI}, GraphAlign ~\cite{graphalign}, and GraphAlign++  ~\cite{graphalign++}  utilize prior knowledge of projection calibration matrices to project point clouds onto the corresponding images via local graph modeling for addressing feature alignment. BEV-based methods  ~\cite{bevfusion-mit, bevfusion-pku, ObjectFusion, song2024graphbev, yin2024isfusion} merge LiDAR and camera representations efficiently into Bird's Eye View (BEV) space. Although the pioneer BEVFusion ~\cite{bevfusion-mit} has demonstrated high performance typically evaluated on pristine datasets like nuScenes, it ignores real-world complexities, particularly the issue of feature misalignment which poses obstacles for its practical application ~\cite{song2024robustness,zhujun_benchmarking}. Therefore, it is essential for future research in multi-modal 3D object detection to address issues like feature misalignment to ensure robust performance in real-world scenarios.

\subsection{Contrastive Learning}
Contrastive learning aims to learn effective representation by pulling semantically close neighbors together and pushing apart non-neighbors ~\cite{hadsell2006dimensionality}. In this paradigm, the model strives to map similar samples to nearby regions in the representation space, while mapping dissimilar samples to distant regions ~\cite{jaiswal2020survey,tian2020makes}.  
Currently, contrastive learning has been extensively studied in Natural Language Processing ~\cite{zhang2022contrastive} and Computer Vision ~\cite{le2020contrastive,clip,trex2,fan2023contrastive,xiao2021region,wang2021dense}. 
Specifically, CLIP ~\cite{clip} leverages contrastive learning for multi-modal pretraining between large-scale text and image data. T-Rex2 ~\cite{trex2} integrates text and visual prompts in object detection models through contrastive learning. WCL ~\cite{fan2023contrastive} incorporates contrastive learning to enhance depth prediction processes. ReSim ~\cite{xiao2021region} learns region representations from different sliding windows of the same image. DenseCL ~\cite{wang2021dense} optimizes pixel-level contrastive loss between two different images. Overall, contrastive learning excels in learning the similarity of cross-modal features and the invariance of single-modal features.

In this work, to address the BEV feature misalignment of LiDAR and camera, we propose a novel multi-modal framework named \textbf{ContrastAlign}. It borrows the idea of contrastive learning to enhance the alignment of heterogeneous modalities. 
        
\section{Is Contrastive Learning All You Need for Feature Alignment?}
In this section, we conduct a thorough analysis of how contrastive learning addresses the alignment issues between multi-modal inputs (LiDAR and Camera) to substantiate the validity of our approach. The following sections detail its mathematical principles and rationale for feature alignment.

Let \( X = \{x_1, x_2, \ldots, x_n\} \) represent a set of samples, where each sample \( x_i \) generates a feature representation \( f(x_i) \). A positive pair \( (x_i, x_j) \) consists of different views of the same object, while a negative pair \( (x_i, x_k) \) comprises different objects.

The objective of contrastive learning is to minimize the following contrastive loss function:

\begin{equation}
L(i,j) = -\log \frac{\exp(\text{sim}(f(x_i), f(x_j))/\tau)}{\sum_{k=1}^{n} \exp(\text{sim}(f(x_i), f(x_k))/\tau)}
\end{equation}

Here, \( \text{sim}(a, b) \) denotes the similarity metric, typically using cosine similarity:

\begin{equation}
\text{sim}(a, b) = \frac{a \cdot b}{\|a\| \|b\|}
\end{equation}

The temperature parameter \( \tau \) controls the smoothness of the distribution, and \( \mathbb{1}_{[k \neq i]} \) ensures negative pairs exclude the current sample.

In the feature alignment process, let \( F_L \) and \( F_C \) represent the features extracted from LiDAR and camera modalities, respectively. The goal is to utilize contrastive learning to bring these features closer in a shared representation space. Specifically, for positive pairs \( (F_L^i, F_C^j) \), we aim to minimize their distance \( d(F_L^i, F_C^j) \), while maximizing the distance for negative pairs \( (F_L^i, F_C^k) \):

\begin{equation}
d(F_L^i, F_C^j) = \|F_L^i - F_C^j\|^2
\end{equation}

This leads to the following objective function:

\begin{equation}
\min_{F_L, F_C} \sum_{(i,j) \in P} d(F_L^i, F_C^j) - \sum_{(i,k) \in N} d(F_L^i, F_C^k)
\end{equation}

where \( P \) denotes the set of positive pairs and \( N \) the set of negative pairs.

\section{Methodology}
To address the feature misalignment issue, we propose a robust fusion framework named \textbf{ContrastAlign} which includes our newly designed C-Instance and L-Instance modules, and a InstanceFusion module. 
The overview of our framework is provided in Figure \ref{fig:framework}. In the following sub-sections, 
we first introduce the general overview of the proposed ContrastAlign. 
Subsequently, we delve into the details of the C-Instance and L-Instance modules. 
After that, 
we elaborate on the crucial design steps of the InstanceFusion module.

\subsection{Overview of ConstrastAlign}
\label{sec:overallframework}
The overall framework, as described and shown in Figure \ref{fig:framework}, mainly consists of four modules: Multi-modal Encoders, C-Instance and L-Instance modules, InstanceFusion, and Detection Head.

\noindent\textbf{Multi-modal Encoders.}
ContrastAlign is built upon the foundation of BEVFusion \cite{bevfusion-mit}. Within the camera branch, we employ the Swim Transformer \cite{swimtransformer} as a feature extractor for multi-camera setups, following LSS \cite{lss} to obtain camera BEV features defined as $F^C_{B} \in  \mathbb{R}^{B_{S} \times C_{C} \times H_B \times W_B}$, where $B_{S}$ represents the batch size, $C_{C}$ denotes the channels of features, and $H_B$, $W_B$ are the height and the width of the features, respectively. In the LiDAR branch, we employ TransFusion-L  \cite{Transfusion} to output LiDAR BEV features defined as $F^L_{B} \in  \mathbb{R}^{B_{S} \times C_{L} \times H_B \times W_B}$, where $C_{L}$ denotes the channels of features.

\noindent\textbf{C-Instance and L-Instance Modules.}
Previous methods like BEVFusion \cite{bevfusion-mit,bevfusion-pku} directly concatenate 
$F^C_{B}$ and $F^L_{B}$. The operation can not distinct instance features and background features in the BEV space. 
Compared to background features, instance features such as cars and pedestrians are more important for 3D detection.
To obtain LiDAR and camera instance features, we introduce C-Instance and L-Instance modules, which predict instance features through score filtering. 

\noindent\textbf{InstanceFusion.}
We introduce the \textbf{InstanceFusion} module to achieve cross-modal alignment during the BEV fusion of LiDAR and camera. This module, detailed in {\bf InstanceFusion} Section, 
represents our core innovation. It leverages the C-Instance and L-Instance modules to provide LiDAR and camera instance features, which are then aligned 
into the BEV space.

\noindent\textbf{Detection Head.}
We follow TransFusion ~\cite{Transfusion} to generate the final 3D detection results. During the training process, we incorporate the InfoNCE loss as described in Eq. (6). 
Meanwhile, we utilize Focal loss  ~\cite{lin2017focal} and L1 loss for classification and 3D bounding box regression, respectively.

\subsection{C-Instance and L-Instance Modules}
\label{Sec:Instance}

\begin{figure*}[t]
\centering
\includegraphics[width=1\textwidth]{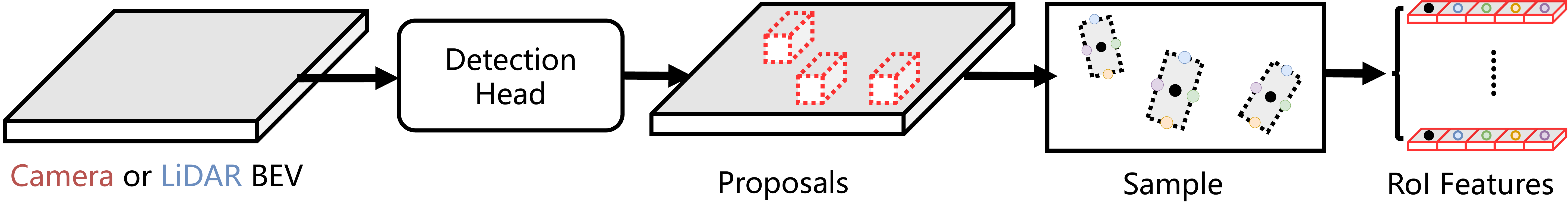}
\caption[ ]{The pipeline of the \textbf{C-Instance} and \textbf{L-Instance} modules. Notably, the Detection Head of \textbf{C-Instance} adopts the CenterPoint  \cite{Centerpoint} Head, while the Detection Head of L-Instance adopts the VoxelNeXt  \cite{chen2023voxenext} Head.
}
\label{fig:instance}
\end{figure*}



As shown in Figure \ref{fig:instance},
C-Instance and L-Instance modules aim to extract instance features from the BEV features of LiDAR and camera. 
Generally, extracting instance features requires point-based operations  \cite{Pointnet}, but these methods are inefficient. We attempt to avoid time-consuming operations like FPS  \cite{Pointrcnn}, and therefore, choose to use an additional detection head to perform this step on the BEV feature map.
For the camera BEV features $F^C_{B}$, we follow BEVDet \cite{huang2021bevdet},  employ CenterPoint Head  \cite{Centerpoint} to generate proposals $P_C$. For the LiDAR BEV features $F^L_{B}$, we employ VoxelNeXt Head  \cite{chen2023voxenext} to generate proposals $P_L$. 
It is worth noting that, influenced by submanifold sparse convolution  \cite{Second,chen2022focalsconv}, we adopt multiple stacked max pool layers for sparse max pooling during the inference of $P_C$, aiming to avoid using NMS.
During inference processing, sparse max pooling  \cite{chen2022focalsconv} is employed to score the selected proposals.
LiDAR and camera proposals, $P_C$ and $P_L$, consist of features including positions $(\delta_x, \delta_y) \in \mathbb{R}^2$, heights $z_{c} \in \mathbb{R}$, dimensions $(w, h, l) \in \mathbb{R}^3$, orientation angles $\alpha \in \mathbb{R}$, and proposal scores 
and labels. 

Subsequently, we employ feature sampling  \cite{Centerpoint} to generate RoI features $F_{C}^{RoI}$ and $F_{L}^{RoI}$ from $F^C_{B}$ and $F^L_{B}$. Specifically, for any proposal $p_c \in P_C$ or $p_l \in P_L$, we select its positions $(\delta_x, \delta_y)$ and sample points at the center of each boundary line of the bounding box. 
Each proposal generates a set of sampling points $[c, c_{\uparrow}, c_{\downarrow}, c_{\leftarrow}, c_{\rightarrow}]$, where $c$ denotes its positions $(\delta_x, \delta_y)$. Apart from $c$, the other sampling points yield sampling results on the LiDAR and camera BEV features using interpolation algorithms  \cite{Centerpoint}. Finally, the LiDAR and camera RoI features, $F_{C}^{RoI} \in \mathbb{R}^{B_S \times N \times 
 (5 \times C_{RoI})}$ and $F_{L}^{RoI} \in \mathbb{R}^{B_S \times M \times (5 \times  C_{RoI})}$, are formed by concatenating all sampling features, where $B_S$ denotes batch size, $N$ and $M$ denote the numbers of LiDAR and camera RoI features, 5 denotes the number of $[c, c_{\uparrow}, c_{\downarrow}, c_{\leftarrow}, c_{\rightarrow}]$, and $C_{RoI}$ denotes the channel of $F_{C}^{RoI}$ and $F_{L}^{RoI}$.

\subsection{InstanceFusion}
\label{Sec:InstanceFusion}

\begin{algorithm}[t]
    \caption{InstanceFusion Workflow} 
    \label{algorithm:ContrastiveLearning}
    \SetAlgoNoLine 
    \KwIn{
    
      ROI features: $F_{C}^{RoI}$ and $F_{L}^{RoI}$  
    
      Position Index: $B_{C}^{RoI}$ and $B_{L}^{RoI}$  
    
      IoU threshold: $\tau = 0.2$ 
    
      Number of positive samples' neighbors: $K = 8$
    }
    \KwOut{
    
      Fusion features $F^{Fuse}$
    }
    
    \While{InstanceFusion}{
      \SetKwProg{Fn}{Function}{}{}
      \Fn{IoU ($B_{C}^{RoI}$, $B_{L}^{RoI}$)}{
        \For{$i=1$ to $N$, $j=1$ to $M$}{
          $IoU_{i,j} = \frac{A_i}{A_u}$
        }
      }
      $I^{pos} = f^{pos}(IoU(B_{C}^{RoI}, B_{L}^{RoI}), \tau)$\\ 
      $F^{pos} = f^F_{pos}(F_{C}^{RoI}, F_{L}^{RoI}, I^{pos})$\\
      \For{positions $q$ of $B_{L}^{RoI}$}{
        \Fn{Shoot ($q$, $K$)}{
          $NBR = \text{KD-Tree}(q, K)$
          $F^{neg} = f^F_{neg}(F_{C}^{RoI}, NBR)$
        }
      }
      \Fn{InfoNCELoss ($F^{pos}$, $F^{neg}$)}{
        \For{each $F^{neg}_i$ in $F^{neg}$}{
          $\mathcal{L}_{align} = - \log \frac{\exp (F^{pos}_L \cdot F^{pos}_C)}{\sum_{k=1}^{K} \exp (F^{pos}_L \cdot F^{neg}_i)}$
        }
      }
    
    
    }
\end{algorithm}

We propose the \textbf{InstanceFusion} module, as shown in Algorithm \ref{algorithm:ContrastiveLearning}, to achieve cross-modal alignment of instance features.
For positive samples, we project LiDAR instance features onto the camera BEV space and consider the matched camera instance features as positive samples. For negative samples, we select the $K$ nearest neighboring instance features around the camera positive samples as negative samples. 
During inference, LiDAR instance features calculate similarity scores with $K$ neighboring camera instances and select the instance feature with the highest score as the matching aligned instance feature. The whole process of InstanceFusion module are as follows.
To construct positive and negative sample pairs for contrastive learning, we establish neighboring instance feature relationships based on LiDAR and camera Instance features using a graph matching approach. During training, LiDAR instance features are projected onto the camera BEV space. Instances, that overlap each other, are directly considered positive samples, while those with close proximity in the KNN graphs  are considered negative samples. 
During the inference process, the LiDAR instance features calculate the similarity score between the $K$ camera instances through graph matching, and the camera instance feature with the highest score is selected as the matching instance.

First, due to the absence of depth errors in the LiDAR BEV feature implementation, we utilize the generated LiDAR RoI features to retrieve the corresponding camera RoI features. 
For any pair of bounding box predictions from $P_C$ with $N$ samples and $P_L$ with $M$ samples, we calculate the Intersection over Union (IoU) by computing the intersection and union areas of their 2D bounding boxes. We define the bounding box sets $\text{B}_{C}^{RoI} = \{ (x_i,y_i,w_i,h_i) | i\in[1, 2...N]\}$ and $\text{B}_{L}^{RoI} = \{ (x_i,y_i,w_i,h_i) | i\in[1, 2...M]\}$  based on the positions $(\delta_x, \delta_y)$ and dimensions $(w, h)$ of proposals. 
To obtain the groups of positive samples for $F^{pos}$, we execute that

\begin{equation}
I^{pos} = f^{pos}(f^{IoU}(\text{B}_{C}^{RoI}, \text{B}_{L}^{RoI}), \tau),
\label{fun:IOU}
\end{equation}
\begin{equation}
F^{pos} = f^F_{pos}(F_{C}^{RoI}, F_{L}^{RoI}, I^{pos}),
\end{equation}
where $I^{pos}\in \mathbb{R}^{N \times M}$ represents the index information, $f^{pos}(\cdot)$ is the function of the index, $\tau$ is the threshold of IoU, $f^{IoU}(\cdot)$ is the function of IoU, $f^F_{pos}(\cdot)$ is the function of the feature index, $F^{pos} \in \mathbb{R}^{B_S \times (2 \times N_{pos}) \times (5 \times C_{RoI})}$ denotes the positive samples result, $N_{Pos}$ denotes the number of positive samples. Nobly, each set of samples from $F^{pos}$ contains an $F_{L}^{pos}$ and an $F_{C}^{pos}$.

After obtaining the positive samples, we proceed to construct negative samples. 
Due to the impact of feature misalignment, offsets occur in a region. We utilize graph matching like the KNN algorithm to construct the neighbors' relations of positive samples and consider each positive sample's neighbor features as negative. Specifically, the construction of negative samples is as follows.
\begin{equation}
I^{neg} = \operatorname{KNN}(I_{i,j}^{pos}, \text{B}_{C}^{RoI}, K),
\label{fun:KNN}
\end{equation}
\begin{equation}
F^{neg} = f^F_{neg}(F_{C}^{RoI}, I^{neg}),
\end{equation}
where $I^{neg} \in \mathbb{R}^{N \times M \times K}$ represents the index information of negative samples, $\operatorname{KNN}(\cdot)$ represents the KNN function that searches for $K$ nearest samples of $I_{i,j}^{pos}$, $f^F_{neg}(\cdot)$ is the function of the feature index, $F^{neg} \in \mathbb{R}^{B_S \times (K \times N_{pos}) \times (5 \times C_{RoI})}$ denotes the negative samples.

To align the heterogeneous modal features in the $K$ neighbor features, we calculate the similarity scores using cosine and select the maximum score to align the RoI features of LiDAR and camera. Where the cosine similarity $\cos (\cdot)$ is defined as 

\begin{equation}
\cos(F^{pos}, F^{neg}) = \frac{F^{pos} \cdot F^{neg}}{\|F^{pos}\| \|F^{neg}\|}.
\end{equation}

The contrastive alignment can be regarded as a mutual distillation process, whereby each modality contributes to and benefits from the exchange of knowledge. During training, to increase alignment of instance features, we use the InfoNCE Loss \cite{infonce} to measure the distance between positive and negative samples, defined as 

\begin{equation}
\mathcal{L}_{\text {align }} =- \log \frac{\exp \left(F_{L}^{pos} \cdot F_{C}^{pos}\right)}{\sum_{i=1}^{K} \exp \left(F_{L}^{pos} \cdot F_i^{neg}\right)},
\end{equation}
where $F_i^{neg}$ represents the $i$-th negative sample of $F^{neg}$.

Finally, positive samples are concatenated with the BEV features along the channel dimensions. By following BEVFusion  \cite{bevfusion-mit}, the BEV features from the two branches are further concatenated and then passed through a convolutional network to extract the cascaded features.

\section{Experiments}
In this section, we present the experimental setup of ContrastAlign and evaluate the performance of 3D object detection on the nuScenes  \cite{nuscenes} dataset. Additionally, we follow the solution provided by Ref.  \cite{zhujun_benchmarking} to simulate feature misalignment scenarios.

\begin{table*}[t]
\scriptsize
\centering
  \caption[ ]{Comparison with the SOTA methods on the nuScenes validation set. `C.V.', `Motor.', `Ped.', and ` T.C.' are short for construction vehicles, motorcycles, pedestrians, and traffic cones.
  }
  \renewcommand\arraystretch{0.9}
  \tabcolsep=1.5mm 
  \resizebox{\linewidth}{!}{
  \begin{tabular}{l|c|cc|ccccc ccccc }
    \toprule
Method       &Modality     & mAP  & NDS & Car  & Truck & C.V. & Bus  & Trailer & Barrier & Motor. & Bike & Ped. & T.C. \\ 
\midrule
TransFusion-L \cite{Transfusion} &L & 65.1 &70.1& 86.5& 59.6& 25.4& 74.4& 42.2& 74.1& 72.1& 56.0& 86.6& 74.1\\
FUTR3D \cite{chen2023futr3d}&LC& 64.2& 68.0& 86.3& 61.5& 26.0& 71.9& 42.1& 64.4& 73.6& 63.3& 82.6& 70.1\\
TransFusion \cite{Transfusion}&LC&  67.3 &71.2& 87.6& 62.0& 27.4& 75.7& 42.8& 73.9& 75.4& 63.1& 87.8& 77.0\\ 
ObjectFusion  \cite{ObjectFusion} &LC&69.8& 72.3 &89.7& 65.6& 32.0& \textbf{77.7}& 42.8& 75.2& 79.4& 65.0& 89.3& 81.1\\
\midrule
BEVFusion  \cite{bevfusion-mit}&LC & 68.5 & 71.4  & 89.2 & 64.6 & 30.4 & 75.4 & 42.5 & 72.0 & 78.5 & 65.3 & 88.2 & 79.5\\
\textbf{ContrastAlign}& LC&  70.3&72.5& 89.5&66.0& 32.9& 76.8& 45.5& 75.6& 79.7& 66.9& 88.8 &81.2\\ 
 &
 &\textit{\fontsize{6}{0}\selectfont\textcolor{red}{+1.8}}
 &\textit{\fontsize{6}{0}\selectfont\textcolor{red}{+1.1}}
 &\textit{\fontsize{6}{0}\selectfont\textcolor{red}{+0.7}}
 &\textit{\fontsize{6}{0}\selectfont\textcolor{red}{+1.4}}
 &\textit{\fontsize{6}{0}\selectfont\textcolor{red}{+2.4}}
 &\textit{\fontsize{6}{0}\selectfont\textcolor{red}{+1.4}}
 &\textit{\fontsize{6}{0}\selectfont\textcolor{red}{+3.0}}
 &\textit{\fontsize{6}{0}\selectfont\textcolor{red}{+3.6}} 
 &\textit{\fontsize{6}{0}\selectfont\textcolor{red}{+1.2}}
 &\textit{\fontsize{6}{0}\selectfont\textcolor{red}{+1.6}}
 &\textit{\fontsize{6}{0}\selectfont\textcolor{red}{+0.6}}
 &\textit{\fontsize{6}{0}\selectfont\textcolor{red}{+1.7}}
 \\
\midrule
GraphBEV  \cite{song2024graphbev}&LC & 70.1& 72.9& 89.9& 64.7& 31.1& 76.0& 43.8& 76.0& 80.1& 67.5& 89.2& 82.2\\
\textbf{ContrastAlign}& LC&  \textbf{71.5}&\textbf{73.7}& \textbf{90.1}& \textbf{67.1}& \textbf{33.1}& 77.2& \textbf{46.4}& \textbf{77.5}& \textbf{81.2}& \textbf{68.4}& \textbf{90.6} &\textbf{83.5}\\ 
 &
 &\textit{\fontsize{6}{0}\selectfont\textcolor{red}{+1.4}}
 &\textit{\fontsize{6}{0}\selectfont\textcolor{red}{+0.8}}
 &\textit{\fontsize{6}{0}\selectfont\textcolor{red}{+0.2}}
 &\textit{\fontsize{6}{0}\selectfont\textcolor{red}{+2.4}}
 &\textit{\fontsize{6}{0}\selectfont\textcolor{red}{+2.0}}
 &\textit{\fontsize{6}{0}\selectfont\textcolor{red}{+1.2}}
 &\textit{\fontsize{6}{0}\selectfont\textcolor{red}{+2.6}}
 &\textit{\fontsize{6}{0}\selectfont\textcolor{red}{+1.5}} 
 &\textit{\fontsize{6}{0}\selectfont\textcolor{red}{+1.1}}
 &\textit{\fontsize{6}{0}\selectfont\textcolor{red}{+0.9}}
 &\textit{\fontsize{6}{0}\selectfont\textcolor{red}{+1.4}}
 &\textit{\fontsize{6}{0}\selectfont\textcolor{red}{+1.3}}
 \\
\bottomrule
\end{tabular} }

\label{tab_nuScens_val_clean}
\end{table*}

\begin{table}[t]
\scriptsize
\centering
\caption{Comparison with SOTAs on the nuScenes val dataset with nosiy setting \cite{zhujun_benchmarking} including  Spatial Misalignment \& Temporal Misalignment. \textcolor{blue}{Blue} denotes clean mAP, \textcolor{red}{Red} denotes noisy mAP, and RCE denotes Relative Corruption Error from Ref.\cite{zhujun_benchmarking}.  }

  \renewcommand\arraystretch{1}
  \tabcolsep=0.8mm 
   \resizebox{\linewidth}{!}{
  \begin{tabular}{l|c|c|c|c|c}
\toprule
           \multirow{2}{*}{Method}           & \multirow{2}{*}{mAP$_{clean}$}  & \multicolumn{2}{c|}{mAP$_{noisy}$}& \multicolumn{2}{c}{RCE ↓ (\%)}\\ 
           \cmidrule(r){3-6} 
           
         & &  Spatial  & Temporal &  Spatial  & Temporal 
           \\
\midrule
Focals Conv-F~\cite{chen2022focalsconv}   & \textcolor{blue}{67.7} & \textcolor{red}{18.7}  & \textcolor{red}{21.3}& \textcolor{black}{72.4}  & \textcolor{black}{68.5}		\\

AutoAlignV2~\cite{autoalignv2}  & \textcolor{blue}{68.5} & \textcolor{red}{62.9}	 & \textcolor{red}{62.8}& \textcolor{black}{8.2}  & \textcolor{black}{8.3}		\\  
\midrule
TransFusion \cite{Transfusion}  & \textcolor{blue}{66.4} & \textcolor{red}{59.2}	 & \textcolor{red}{43.7}&  \textcolor{black}{10.8}  & \textcolor{black}{34.2}			\\

FUTR3D \cite{chen2023futr3d} & \textcolor{blue}{64.2} & \textcolor{red}{58.8}	 & \textcolor{red}{51.4}	& \textcolor{black}{8.4}  & \textcolor{black}{19.9}	\\

DeepInteraction \cite{DeepInteraction} & \textcolor{blue}{69.9} & \textcolor{red}{62.1}	 & \textcolor{red}{59.4}	& \textcolor{black}{11.1}  & \textcolor{black}{15.0}	\\ 

CMT \cite{cmt}& \textcolor{blue}{70.3} & \textcolor{red}{65.2}	 & \textcolor{red}{58.7}	& \textcolor{black}{7.2}  & \textcolor{black}{16.5}	\\

\midrule
BEVFusion \cite{bevfusion-mit} & \textcolor{blue}{68.5} & \textcolor{red}{60.8}	 & \textcolor{red}{49.0}	& \textcolor{black}{11.2}  & \textcolor{black}{28.5}	\\ 
\textbf{ContrastAlign}     & \textcolor{blue}{\textbf{70.3}} & \textcolor{red}{\textbf{68.1}}	&  \textcolor{red}{\textbf{63.1}}& 3.1  & \textcolor{black}{10.2}	\\ 
\midrule
GraphBEV  \cite{song2024graphbev}& \textcolor{blue}{70.1} & \textcolor{red}{69.1}	 & \textcolor{red}{55.1}	& \textcolor{black}{1.4}  & 21.4	\\ 
\textbf{ContrastAlign}     & \textcolor{blue}{\textbf{71.5}} & \textcolor{red}{\textbf{70.5}}	&  \textcolor{red}{\textbf{66.2}}& \textbf{1.4}  & \textcolor{black}{\textbf{7.4}}	\\

\bottomrule
\end{tabular}}
\label{tab_nuScenes_temporal_misalignment}
\end{table}

\begin{table*}[t]
\scriptsize
\centering
  \caption{Comparison with the SOTA methods on the nuScenes  \textbf{test}  set. `C.V.', `Motor.', `Ped.', and `T.C.' are short for construction vehicle, motorcycle, pedestrian, and traffic cone, respectively. `L' means only LiDAR data are used, `LC' denotes the use of both LiDAR and camera data. 
  The best performances are marked in \textbf{bold}.}
  \renewcommand\arraystretch{0.9}
  \tabcolsep=1.99mm 
  
  \resizebox{\linewidth}{!}{
  \begin{tabular}{l|c|cc|ccccc ccccc }
    \toprule
Method      &Modality       & mAP  & NDS  & Car  & Truck & C.V. & Bus  & Trailer & Barrier & Motor. & Bike & Ped. & T.C. \\ 
\midrule
VoxelNeXt  \cite{chen2023voxenext} & L& 64.5& 70.0 &84.6& 53.0& 28.7& 64.7 &55.8& 74.6& 73.2& 45.7& 85.8& 79.0\\
TransFusion-L \cite{Transfusion}& L& 65.5& 70.2 &86.2& 56.7& 28.2& 66.3& 58.8& 78.2& 68.3& 44.2& 86.1& 82.0\\ 
\midrule
GraphAlign  \cite{graphalign}& LC  &66.5&70.6& 87.6& 57.7& 26.1& 66.2& 57.8& 74.1& 72.5& 49.0& 87.2 &86.3\\
AutoAlignV2 \cite{autoalignv2}& LC & 68.4 & 72.4 & 87.0 & 59.0 & 33.1 & 69.3 & 59.3 & - & 72.9 & 52.1 & 87.6 & -\\
UVTR \cite{uvtr} & LC &67.1& 71.1& 87.5& 56.0& 33.8 &67.5 &59.5& 73.0& 73.4& 54.8 &86.3& 79.6\\
TransFusion  \cite{Transfusion}& LC & 68.9 & 71.7 & 87.1 & 60.0 & 33.1 & 68.3 & 60.8 & 78.1 & 73.6 & 52.9 & 88.4 & 86.7\\
DeepInteraction \cite{DeepInteraction} &LC  &70.8& 73.4& 87.9& 60.2& 37.5& 70.8& 63.8& 80.4& 75.4& 54.5& 90.3& 87.0 \\
ObjectFusion  \cite{ObjectFusion}& LC &71.0& 73.3& 89.4& 59.0 & 40.5 & 71.8 & 63.1& 76.6 & 78.1 & 53.2 & 90.7 & 87.7\\
UniTR \cite{wang2023unitr} &LC& 70.9& 74.5 &87.9& 60.2& 39.2& 72.2& 65.1& 76.8 &75.8 &52.2 &89.4& 89.7\\
FocalFormer3D \cite{focalformer}& LC& 71.6& 73.9 &88.5& 61.4& 35.9& 71.7& \textbf{66.4}& 79.3 &\textbf{80.3}& 57.1& 89.7& 85.3\\
MSMDFusion\cite{msmdfusion}& LC& 71.5&74.0& 88.4& 61.0& 35.2& 71.4& 64.2& 80.7& 76.9 &58.3& 90.6& 88.1\\
SparseFusion\cite{sparsefusion}& LC&72.0 &73.8& 88.0& 60.2& 38.7& 72.0& 64.9& 79.2& 78.5& 59.8& 90.9& 87.9\\
CMT\cite{cmt}&LC&72.0& 74.1& 88.0& \textbf{63.3}& 37.3& \textbf{75.4}& 65.4& 78.2& 79.1& \textbf{60.6} &87.9 &84.7\\
\midrule
BEVFusion  \cite{bevfusion-mit}& LC & 70.2 & 72.9 & 88.6 & 60.1 & 39.3& 69.8 & 63.8 & 80.0 & 74.1 & 51.0 & 89.2 & 86.5\\ 
\textbf{ContrastAlign}& LC& 72.2 & 73.8& 89.0& 60.9 & 41.1 & 73.1 & 64.6& 82.2 & 75.9 & 54.5 & 90.9& 89.3\\
&
&\textit{\fontsize{6}{0}\selectfont\textcolor{red}{+2.0}}
&\textit{\fontsize{6}{0}\selectfont\textcolor{red}{+0.9}} 
&\textit{\fontsize{6}{0}\selectfont\textcolor{red}{+0.4}}
&\textit{\fontsize{6}{0}\selectfont\textcolor{red}{+0.8}}
&\textit{\fontsize{6}{0}\selectfont\textcolor{red}{+1.8}}
&\textit{\fontsize{6}{0}\selectfont\textcolor{red}{+4.3}}
&\textit{\fontsize{6}{0}\selectfont\textcolor{red}{+0.8}}
&\textit{\fontsize{6}{0}\selectfont\textcolor{red}{+2.2}}
&\textit{\fontsize{6}{0}\selectfont\textcolor{red}{+1.8}}
&\textit{\fontsize{6}{0}\selectfont\textcolor{red}{+3.5}}
&\textit{\fontsize{6}{0}\selectfont\textcolor{red}{+1.7}}
&\textit{\fontsize{6}{0}\selectfont\textcolor{red}{+2.8}}
\\
\midrule
GraphBEV  \cite{song2024graphbev}& LC & 71.7& 73.6 &89.2 &60.0& 40.8& 72.1& 64.5& 80.1& 76.8& 53.3& 90.9& 88.9\\ 
\textbf{ContrastAlign}& LC& \textbf{72.9} & \textbf{74.9}& \textbf{89.5}& 61.5 & \textbf{41.8} & 74.5 & 65.7& \textbf{82.6} & 77.4 & 54.8 & \textbf{91.6}&  \textbf{90.0}\\
&
&\textit{\fontsize{6}{0}\selectfont\textcolor{red}{+1.2}}
&\textit{\fontsize{6}{0}\selectfont\textcolor{red}{+1.3}} 
&\textit{\fontsize{6}{0}\selectfont\textcolor{red}{+0.3}}
&\textit{\fontsize{6}{0}\selectfont\textcolor{red}{+1.5}}
&\textit{\fontsize{6}{0}\selectfont\textcolor{red}{+1.0}}
&\textit{\fontsize{6}{0}\selectfont\textcolor{red}{+2.4}}
&\textit{\fontsize{6}{0}\selectfont\textcolor{red}{+1.2}}
&\textit{\fontsize{6}{0}\selectfont\textcolor{red}{+2.5}}
&\textit{\fontsize{6}{0}\selectfont\textcolor{red}{+0.6}}
&\textit{\fontsize{6}{0}\selectfont\textcolor{red}{+1.5}}
&\textit{\fontsize{6}{0}\selectfont\textcolor{red}{+0.7}}
&\textit{\fontsize{6}{0}\selectfont\textcolor{red}{+1.1}}
\\
\bottomrule
\end{tabular} }
\label{tab_nuScens_test}
\end{table*}

\begin{table*}[t]
\scriptsize
\centering
  \caption{Comparison with prior methods on Argoverse2 validation set. Metrics: mAP (\%)↑ for the overall results, AP (\%)↑ for each category. * denotes result from Ref. \cite{chen2023voxenext}. $^{\dagger}$ denotes result re-implentment.}
  \renewcommand\arraystretch{0.9}
  \tabcolsep=0.5mm 
  
  \resizebox{\linewidth}{!}{
  \begin{tabular}{l|c|ccccc ccccc ccccc ccccc }
    \toprule
Method      &mAP& Veh.& Bus& Ped.& Stop.& Box.& Boll.& C-B.& M.-list& MPC.& M.-cycle& Bicycle& A-B.& School.& Truck.& C-C.& V-T.& Sign& Large.& Str.& Bic.-list \\ 
\midrule
CenterPoint* \cite{Centerpoint}&22.0& 67.6& 38.9& 46.5& 16.9& 37.4& 40.1& 32.2& 28.6& 27.4& 33.4& 24.5& 8.7& 25.8& 22.6& 29.5& 22.4& 6.3& 3.9& 0.5& 20.1\\
FSD* \cite{fsd}& 28.2& 68.1& 40.9 &59.0& 29.0& 38.5& 41.8& 42.6& 39.7& 26.2& 49.0& 38.6& 20.4& 30.5& 14.8& 41.2& 26.9& 11.9& 5.9 &13.8 &33.4\\
VoxelNeXt* \cite{chen2023voxenext}& 30.0& 71.7& 39.2& 63.1& 39.2& 40.0& 52.5& 63.7& 42.2& 34.9& 42.7& 40.1& 20.1& 25.2& 16.9& 45.7& 22.3& 15.8& 5.9& 9.8& 33.5\\
\midrule
BEVFusion$^{\dagger}$ \cite{bevfusion-mit}& 43.1&83.1& 48.1& 64.2& 48.2& 54.5& 58.1& 65.3& 45.1& 39.8& 49.3& 48.1& 33.1& 38.1& 28.5& 54.1& 37.4& 34.6& 31.2& 34.4& 38.1\\
\textbf{ContrastAlign}& 45.7& 85.0& 51.1& 67.0& 51.7& 57.0& 61.2& 68.0& 48.0& 42.5& 53.5 &52.0& 35.5& 42.3& 31.2& 57.0& 41.0& 36.5 &35.3& 36.8 &41.2\\
\midrule
GraphBEV$^{\dagger}$ \cite{song2024graphbev}& 45.1&84.7&51.6&66.5&51.9&57.4&60.8&67.2&48.3&43.0&53.1&52.5&35.2&42.0&30.8&56.8&41.3&36.0&35.5&36.6&41.0\\
\textbf{ContrastAlign}& \textbf{47.1}&  \textbf{86.0}& \textbf{52.4}& \textbf{68.5}& \textbf{53.2}& \textbf{59.7}& \textbf{62.0}& \textbf{69.5}& \textbf{49.0}& \textbf{43.6}& \textbf{54.0}& \textbf{53.5}& \textbf{37.5}& 
\textbf{43.5}& \textbf{32.0}& \textbf{58.0}& \textbf{41.5}& \textbf{38.0}& \textbf{37.0}& \textbf{37.5}& \textbf{42.0}
\\
\bottomrule
\end{tabular} }
\label{tab_Argoverse2_val}
\end{table*}

\begin{table*}[t]
\scriptsize
\centering
  \caption{Robustness to weather conditions, different ego distances, different sizes on nuScenes  \cite{nuscenes} clean validation set. The evaluation metric is mAP (\%). }
  \renewcommand\arraystretch{0.9}
  \tabcolsep=2.8mm 
  \resizebox{\linewidth}{!}{
  \begin{tabular}{l|ll|lll| lll }
    \toprule
  \multirow{2}{*}{Method} &\multicolumn{2}{c|}{  Different Times }  &\multicolumn{3}{c|}{Different Ego Distances}&\multicolumn{3}{c}{Different Object Sizes}  \\
            & Day  & Night & Near & Middle  & Far & Small & Moderate & Large  \\ 
\midrule

BEVFusion  \cite{bevfusion-mit}&  68.5& 42.8 & 79.4& 64.9& 40.0 & 50.3 &58.7 &64.0 \\ 
\textbf{ContrastAlign}  & 69.8\textit{\fontsize{6}{0}\selectfont\textcolor{red}{+1.3}}& 44.1 \textit{\fontsize{6}{0}\selectfont\textcolor{red}{+1.3}}&79.6 \textit{\fontsize{6}{0}\selectfont\textcolor{red}{+0.2}} & 65.6 \textit{\fontsize{6}{0}\selectfont\textcolor{red}{+0.7}} & 43.9 \textit{\fontsize{6}{0}\selectfont\textcolor{red}{+3.9}}&53.1 \textit{\fontsize{6}{0}\selectfont\textcolor{red}{+3.1}}& 60.3 \textit{\fontsize{6}{0}\selectfont\textcolor{red}{+1.6}}& 64.5 \textit{\fontsize{6}{0}\selectfont\textcolor{red}{+0.5}}\\
\midrule
GraphBEV  \cite{song2024graphbev}&  69.7& 45.1 & 78.6& 65.3& 42.1 & 55.4& 58.3& 63.1 \\ 
\textbf{ContrastAlign}  & 70.3\textit{\fontsize{6}{0}\selectfont\textcolor{red}{+0.6}}& 47.0 \textit{\fontsize{6}{0}\selectfont\textcolor{red}{+1.9}}&79.9 \textit{\fontsize{6}{0}\selectfont\textcolor{red}{+1.3}} & 66.1 \textit{\fontsize{6}{0}\selectfont\textcolor{red}{+0.8}} & 44.6 \textit{\fontsize{6}{0}\selectfont\textcolor{red}{+1.5}}&58.3 \textit{\fontsize{6}{0}\selectfont\textcolor{red}{+2.9}}& 60.2 \textit{\fontsize{6}{0}\selectfont\textcolor{red}{+1.9}}& 64.6 \textit{\fontsize{6}{0}\selectfont\textcolor{red}{+1.5}}\\ 


\bottomrule
\end{tabular} }

\label{tab:nuscenes_robustness}
\end{table*}

\begin{table*}[t]
\scriptsize
\centering
  \caption{Roles of different modules in ContrastAlign for feature alignment on nuScenes validation set under  \textcolor{red}{noisy} misalignment setting. `C.V.', `Motor.', `Ped.', `T.C.', `C.L.', `I.F.', and `LT (ms)' are short for construction vehicle, motorcycle, pedestrian, traffic cone, C-Instance and L-Instance modules, the InstanceFusion module, and latency respectively. All latency measurements are conducted on the same workstation with an A100 GPU.
  }
  \renewcommand\arraystretch{0.9}
  \tabcolsep=2.2mm 
  \resizebox{\linewidth}{!}{
  \begin{tabular}{ccc |ccc|ccccc ccccc }
    \toprule

BEVFusion &C.L.&I.F.          & mAP  & NDS & LT (ms)  & Car  & Truck & C.V. & Bus  & Trailer & Barrier & Motor. & Bike & Ped. & T.C. \\ 

\midrule
$\surd $&&&  60.8 & 65.7 &132.9 & 83.1 & 50.3 & 26.5& 66.4 & 38.0 & 65.0 & 64.9 & 52.8 & 86.1 & 75.1\\

$\surd $&$\surd $&&   61.2&66.1&139.2&  83.0 & 51.3 & 27.1& 66.0 & 38.4 & 65.1 & 65.2 & 53.1 & 86.9 & 76.1\\ 
$\surd $&$\surd $&$\surd $&    68.1 & 70.9&154.4  & 88.6 & 63.9 & 29.3 & 74.5 & 41.8 & 71.6 & 77.9 & 64.9 & 88.6 & 80.0\\  
\bottomrule
\end{tabular} }
\label{tab_nuScens_ablation}
\end{table*}


\begin{table*}[t]
\centering
\caption{The impact of hyperparameter $\tau$ in InstanceFusion module.}

\centering
\scriptsize

\tabcolsep=2.5mm
\renewcommand\arraystretch{0.9}
  \resizebox{\linewidth}{!}{
\begin{tabular}{c|ccc| ccccc ccccc}
\toprule
$\tau$  & mAP & NDS & LT (ms) &Car& Truck& C.V.& Bus& Trailer& Barrier& Motor.& Bike &Ped.& T.C. \\
\midrule
0.05 & 67.3 & 69.2 & 156.3 & 88.1&63.5&29.8&74.0&41.3&71.1&77.4&64.4&88.1&79.5 \\
0.1 & \textbf{68.1} & \textbf{70.9} & 154.4& 88.6 & 63.9 & 29.3 & 74.5 & 41.8 & 71.6 & 77.9 & 64.9 & 88.6 & 80.0 \\
0.2 & 65.0 & 67.3 & 151.2& 84.7 & 60.5 & 24.6 & 71.8 & 39.8 & 69.4 & 75.2 & 62.5 & 85.0 & 76.6 \\
\bottomrule
\end{tabular}}
\label{tab_nus_t}

\end{table*}

\begin{table*}[t]
\centering
\caption{The impact of hyperparameter $K$ in InstanceFusion module.}

\centering
\scriptsize

\tabcolsep=2.6mm
\renewcommand\arraystretch{0.9}
  \resizebox{\linewidth}{!}{
\begin{tabular}{c|ccc | ccccc ccccc}
\toprule
$K$  & mAP & NDS & LT (ms) &Car& Truck& C.V.& Bus& Trailer& Barrier& Motor.& Bike &Ped.& T.C. \\
\midrule
5 & 67.8&  69.7&  151.3&87.5&62.5&28.4&73.5&40.6&70.3&75.9&63.2&86.9&78.9 \\
8 & \textbf{68.1} & \textbf{70.9} & 154.4& 88.6 & 63.9 & 29.3 & 74.5 & 41.8 & 71.6 & 77.9 & 64.9 & 88.6 & 80.0 \\
16& 67.2& 68.1 &159.8& 87.1 & 61.9 & 28.0 & 73.2 & 39.9 & 69.8 & 75.5 & 62.6 & 85.8 & 78.2 \\
\bottomrule
\end{tabular}}
\label{tab_nus_k}

\end{table*}

\subsection{Experimental Setup}
\subsubsection{NuScenes Dataset and Evaluation Metrics}
We evaluate the 3D object detection performance of the proposed ContrastAlign by comparing it with other state-of-the-art approaches on the nuScenes benchmark  \cite{nuscenes}, which is collected with a 32-beam LiDAR and 6 cameras. The six images cover a 360-degree surrounding, and the dataset provides calibration matrices that enable precise projection from 3D points to 2D pixels. It requires detecting 10 object categories that are commonly observed in driving scenarios. We use mAP and NDS across all categories as the primary metrics for evaluation following  \cite{Transfusion, bevfusion-mit, bevfusion-pku,song2024graphbev}. Note that the NDS metric is a weighted average of mAP and other breakdown metrics (e.g., translation, scale, orientation, velocity, and attributes errors). 

In addition, to validate the robustness of \textbf{feature alignment}, we follow \cite{zhujun_benchmarking} to simulate the misalignment between LiDAR and camera projection. It is worth noting that \cite{zhujun_benchmarking} only adds noise to the validation dataset, not to the training and testing datasets.

\subsubsection{Argoverse 2 (AV2) Dataset and Evaluation Metrics}
We further conduct long-range experiments on the recently released Argoverse 2 dataset \cite{Argoverse2} to demonstrate the superiority of our ContrastAlign in long-range detection. AV2 has a large scale dataset, and it contains 1000 sequences in total, 700 for training, 150 for validation, and 150 for testing. In addition to average precision (AP), AV2 adopts a composite score as an evaluation metric, which takes both AP and localization errors into account. The perception range in AV2 is 200 meters (cover area of 400m × 400m), which is much larger than Waymo. Such a large perception range leads to a huge memory footprint for dense detectors.

\subsubsection{Dataset and Evaluation Metrics.}
We conducted experiments on the popular  nuScenes\cite{nuscenes}, and Argoverse2 \cite{Argoverse2} datasets to validate the effectiveness of our approach. The detection ranges of the two datasets are 54 and 200 meters, respectively.  For object detection on the nuScenes dataset, evaluation metrics include mAP and the nuScenes detection score (NDS). mAP is calculated by averaging over the distance thresholds of 0.5m, 1m, 2m, and 4m across all categories. NDS is a weighted average of mAP and five other true positive metrics that measure translation, scaling, orientation, velocity, and attribute errors. For object detection on the Argoverse2 dataset, mAP is adopted as the evaluation metric. We are testing on the AV2 dataset due to its extensive range of long-distance scenarios, where the issue of feature misalignment intensifies with increasing distance. In addition, to validate the robustness of \textbf{feature alignment}, we follow \cite{zhujun_benchmarking} in nuScenes dataset to simulate the misalignment between LiDAR and camera projection. It is worth noting that \cite{zhujun_benchmarking} only adds noise to the validation dataset, not to the training and testing datasets.

\subsubsection{Implementation Details.}

We implement ContrastAlign within PyTorch \cite{paszke2019pytorch}, which is built upon the open-source OpenPCDet \cite{openpcdet}, where our baselines are BEVFusion \cite{bevfusion-mit} and GraphBEV \cite{song2024graphbev}.
In the LiDAR branch, feature encoding is performed by using TransFusion-L  \cite{Transfusion} to obtain LiDAR BEV features, where voxel dimensions are set to [0.075m, 0.075m, 0.2m] and point cloud ranges are specified as [-54m, -54m, -5m, 54m, 54m, 3m] along the X, Y, and Z axes, respectively.
The camera branch employs a Swin Transformer  \cite{swintransformer} as the camera backbone, integrating heads of numbers 3, 6, 12, and 24, and utilizes FPN  \cite{maskrcnn} for fusing multi-scale feature maps. 
The resolution of input images is adjusted and cropped to 256 $\times$ 704.
In the LSS  \cite{lss} configuration, frustum ranges are set with X coordinates [-54m, 54m, 0.3m], Y coordinates [-54m, 54m, 0.3m], Z coordinates [-10m, 10m, 20m], and depth ranges are set to [1m, 60m, 0.5m].
During training, we apply data augmentation for 10 epochs, which includes random flips, rotations (within the range [$-\frac{\pi}{4}$, $\frac{\pi}{4}$]), translations (with std=0.5), and scaling in the range of 0.9 to 1.1 for LiDAR data enhancement. We utilize CBGS  \cite{CBGS} to resample the training data.
Additionally, we use random rotation in [$-5.4^\circ$, $5.4^\circ$] and random resizing in [0.38, 0.55] to augment the images. The Adam optimizer  \cite{adam} is used with a one-cycle learning rate policy, setting the maximum learning rate to 0.001 and weight decay to 0.01. 
The batch size is 24, and training is conducted on 8 NVIDIA GeForce RTX 3090 24G GPUs. During inference, we remove Test Time Augmentation (TTA) data augmentation, and the batch size is set to 1 on an A100 GPU. All latency measurements are taken on the same workstation with an A100 GPU.

\subsection{Main Results}
\noindent\textbf{Results on NuScenes Val set.}
As shown in Table \ref{tab_nuScens_val_clean}, we compare our ContrastAlign with SOTA methods on the nuScenes val set under clean and noisy misalignment settings. Compared to SOTA methods ~\cite{chen2023futr3d,Transfusion,ObjectFusion}, our method achieves the best performance, especially in challenging categories like construction vehicles, trailers, barriers, bikes, and traffic cones, which mainly consist of small and difficult-to-detect objects. On the clean nuScenes val set, our method achieves SOTA performance with 70.3\% mAP and 72.5\% NDS, outperforming the baseline BEVFusion ~\cite{bevfusion-mit} by 1.8\% mAP and 1.1\% NDS, and achieving 71.5\% mAP and 73.7\% NDS in the baseline GraphBEV ~\cite{song2024graphbev}, surpassing it by 1.4\% mAP and 0.8\% NDS. 

\noindent\textbf{Results on MisAlignment in NuScenes Val Set.}
As shown in Table \ref{tab_nuScenes_temporal_misalignment}, our main focus is not on the clean benchmark but on addressing the noisy misalignment issue. In comparison to TransFusion ~\cite{Transfusion} and BEVFusion ~\cite{bevfusion-mit} on the nuScenes val set under noisy misalignment setting, our ContrastAlign achieves SOTA performance. Compared to the baselines BEVFusion ~\cite{bevfusion-mit} and GraphBEV ~\cite{song2024graphbev}, our ContrastAlign improves performance by 7.3\% and 14.1\% mAP for spatial and temporal misalignment in BEVFusion, and by 1.4\% and 11.1\% mAP in GraphBEV. This demonstrates that our method significantly alleviates the misalignment issue.It is worth noting that although SOTAs like CMT ~\cite{cmt} perform well in spatial misalignment, they are not effective in solving temporal misalignment, which is because they only involve feature queries, while we consider the essence of instance alignment.
It is worth noting that TransFusion ~\cite{Transfusion}, as a feature-level method, addresses feature misalignment through attention. However, the global query scheme of TransFusion cannot be directly applied to BEVFusion  ~\cite{bevfusion-mit,bevfusion-pku} due to differences in their architectures or objectives. 
In summary, our ContrastAlign has shown significant improvement not only in the clean setting but also in the noise setting, proving th at the proposed method enables to deal with the misalignment issue.

\noindent\textbf{Results on NuScenes Test Set.}
As shown in Table \ref{tab_nuScens_test}, ContrastAlign achieves 72.9\% mAP and 74.9\% NDS on the nuScenes test benchmark, outperforming GraphBEV  ~\cite{song2024graphbev} by 1.2\% mAP and 1.3\% NDS. And ContrastAlign outperforms BEVFusion~\cite{bevfusion-mit} by by 2.0\% mAP and 0.9\% NDS.
Compared to both LiDAR-based methods such as 
VoxelNeXt ~\cite{chen2023voxenext}, TransFusion-L ~\cite{Transfusion}, and multi-modal methods such as TransFusion ~\cite{Transfusion}, DeepInteraction ~\cite{DeepInteraction}, and ObjectFusion ~\cite{ObjectFusion}, our method achieves SOTA performance. Specifically, compared to ObjectFusion ~\cite{ObjectFusion}, as a solution to BEV feature misalignment, our method outperforms ObjectFusion by 0.8\% mAP and 0.5\% NDS.
What's more, ContrastAlign has significant improvements in construction vehicles, barriers, motorcycles, bikes, pedestrians, and traffic cones, respectively. Although the nuScenes is an open-source clean dataset, it inevitably has minimal feature misalignment issues. 
Our notable improvements on the nuScenes test set indicate that our method effectively mitigates feature misalignment.

\noindent\textbf{Results on Argoverse2 Val Set.}
To validate the effectiveness of our ContrastAlign on the long-range detection, we conducted experiments on the Argoverse2 dataset with a perception range of 200 meters as shown in Table \ref{tab_Argoverse2_val}.
It is worth noting that the proposed ContrastAlign compared with the orther methods including CenterPoint~\cite{Centerpoint}, FSD~\cite{fsd}, VoxelNeXt~\cite{chen2023voxenext}, BEVFusion~\cite{bevfusion-mit} and GraphBEV~\cite{song2024graphbev}, also outperforms the best. Specifically, our ContrastAlign achieves a 2.0\% mAP improvement over GraphBEV, highlighting the method's increased effectiveness with greater distances and more severe feature misalignment.
This indicates that the farther the distance and the more severe the feature misalignment, the more effective ContrastAlign is.

\subsection{Robustness Analysis}
To demonstrate the robustness of our method, we present the results in Table \ref{tab:nuscenes_robustness} under different times, ego distances, and object sizes. Since different times pose challenges to the detection performance of the model, we follow the procedure of BEVFusion ~\cite{bevfusion-mit} and split the val set into Day/Night based on the keywords `day' and `night' in the scene descriptions. Compared to BEVFusion  ~\cite{bevfusion-mit}, our method significantly improves Day and Night scenarios through contrastive learning for feature alignment. Specifically, for night scenarios, our method outperforms BEVFusion by 1.3\%, where depth estimation presents a greater challenge for BEVFusion due to poor lighting conditions.

Far-small objects are more susceptible to feature misalignment. Following BEVFusion ~\cite{bevfusion-mit}, we categorize annotation and prediction ego distances into three groups, Near (0-20m), Middle (20-30m) and Far ($>$30m), summarize the size distributions for each category, defining three equal-proportion size levels, Small, Moderate and Large.
For the different ego distance, we have seen improvements in Near, Middle, and Far, especially in Far, which has increased by 3.9\%. 
This indicates that the farther the distance, the more severe the feature misalignment and the more effective ContrastAlign is.
In the different object sizes, we have seen significant improvements in Small, Moderate, and Large, especially in Small objects, which has increased by 3.1\%.
The improvement of the `Far' and `Small' metrics emphasizes the improvement of our method, which makes our method be visually meaningful since better feature alignment allows for better detection of smaller and farther objects.

\subsection{Ablation Studies}
\noindent\textbf{Roles of different modules in ContrastAlign.}
To analyze the impact of different modules in ContrastAlign on feature misalignment, we conduct experiments on nuScenes validation set under the noise misalignment setting, as shown in Table \ref{tab_nuScens_ablation}. 
We have added C-Instance and L-Instance modules based on the baseline (BEVFusion), simply extracting features from them through additional detection heads such as CenterPoint head and VoxelNeXt head, and then adding them to existing BEV features. We extract RoI features of LiDAR and camera through C-Instance and L-Instance modules and add InstanceFusion to solve the misalignment problem through contrastive learning. 
It can be seen that after adding InstanceFusion module, there is a significant improvement, with mAP increasing from 61.2\% to 68.1\% (an increase of 6.9\%), which is a very significant performance improvement. Meanwhile, the impact on latency is within a controllable range. Evidence demonstrates the effectiveness of incorporating contrastive learning into multi-modal 3D object detection to address feature misalignment issues in BEV features.

\noindent\textbf{Effect of the Hyperparameters $K$ and $\tau$ in InstanceFusion module.}
To assess the influence of various hyperparameters within the InstanceFusion module, an examination of the variables $\tau$ and $K$ was conducted on nuScenes val set under noisy misalignment setting. Where $\tau$ comes from Eq.(\ref{fun:IOU}), which is the threshold of IoU affecting the number of positive sample pairs, and $K$ comes from Eq.(\ref{fun:KNN}), affecting the number of negative samples. Due to the different positive and negative sample pairs at different frames, we cannot provide a fixed number. As shown in Table \ref{tab_nus_t}, when $\tau$ changes, $K$ is fixed at 8 to obtain better performance indicators. As shown in Table \ref{tab_nus_k}, when $K$ changes, $\tau$ is fixed at 0.1. We choose several representative results for analysis. The larger the $\tau$, the smaller the number of positive samples. It can be observed that when $\tau$ is 0.1, the performance is the best, reaching 68.1\% mAP and 70.9\% NDS. The larger $K$, the larger the number of negative samples, and the best performance is achieved when $K$  is 8.

\section{Conclusion}
In this work, we propose a robust fusion framework, ContrastAlign, to address the feature misalignment in BEV-based methods  ~\cite{bevfusion-mit, bevfusion-pku}. Our ContrastAlign utilizes contrastive learning to enhance the alignment of heterogeneous modalities, thereby improving the robustness of LiDAR-camera BEV feature fusion. Specifically, we present the C-Instance and L-Instance modules, which predict camera and LiDAR instance features based on camera and LiDAR BEV features. We propose the InstanceFusion module, which utilizes contrastive learning to generate similar instance features across heterogeneous modalities. During inference, the aligned features with high similarity in neighboring camera instance features are chosen as alignment features to achieve BEV feature alignment. Extensive experiments have demonstrated our method can address the misalignment issue, with significant performance improvement in the misaligned noisy setting ~\cite{zhujun_benchmarking}.

\noindent\textbf{Limitation and Future Work.}
As for limitations, our method can only adapt to LSS-based methods like BEVFusion  \cite{bevfusion-mit, bevfusion-pku} rather than query-based methods like FUTR3D  \cite{chen2023futr3d}. Currently, the calibration matrix of an autonomous driving dataset relies on manual input, inevitably introducing misalignment errors. In the future, we will explore more practical solutions to address the feature misalignment issue by leveraging visual foundational models such as Depth Anything  \cite{depthanything}.

\addtolength{\textheight}{-0cm}



\bibliographystyle{IEEEtran}
\bibliography{IEEEabrv,IROS}\ 
%
%
%

\begin{IEEEbiography}[{\includegraphics[width=1in,height=1.25in,clip,keepaspectratio]{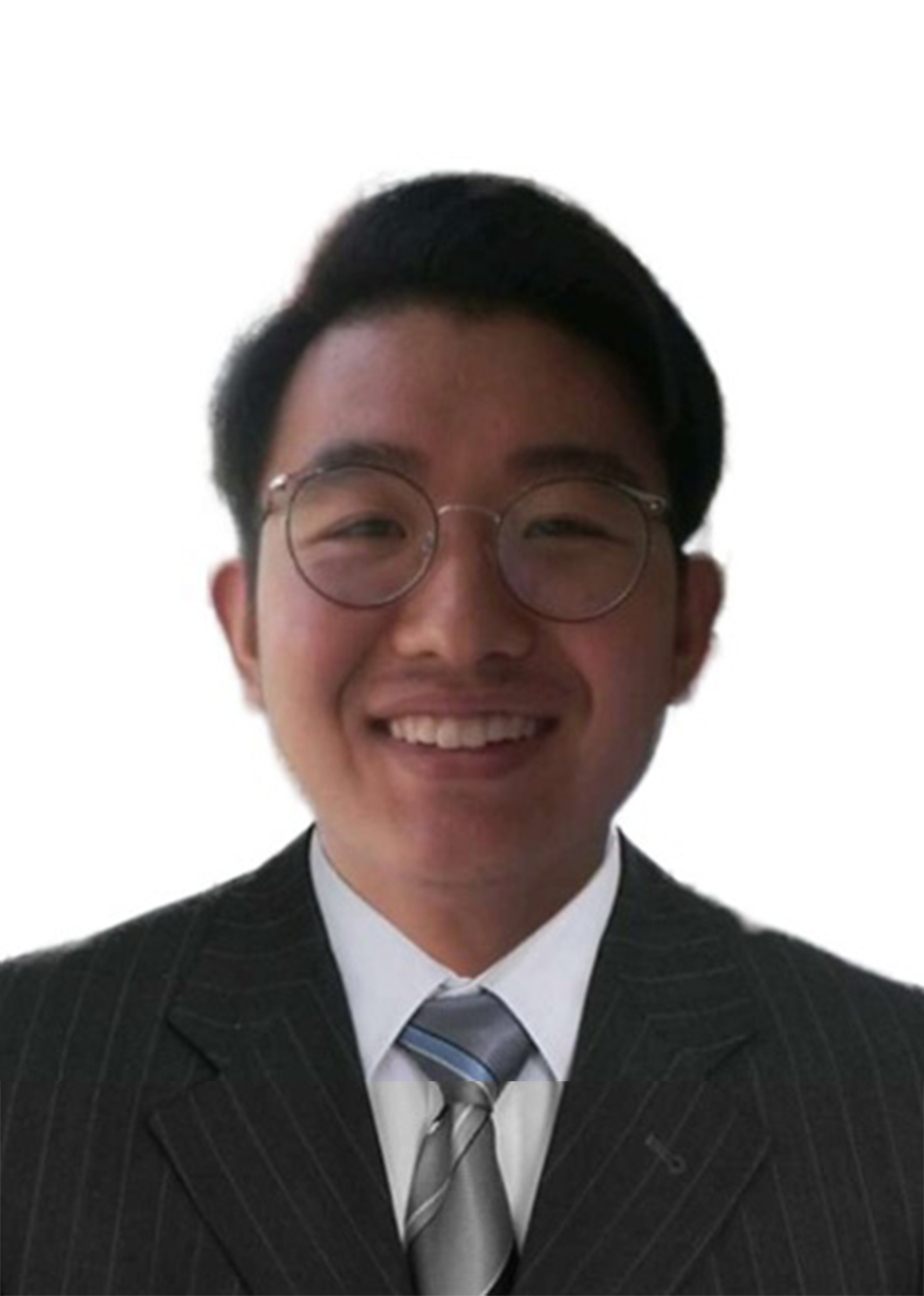}}]{Ziying Song}, was born in Xingtai, Hebei Province, China in 1997. He received the B.S. degree from Hebei Normal University of Science and Technology (China) in 2019. He received a master's degree major in Hebei 
University of Science and Technology (China) in 2022. He is now a PhD student majoring in Computer Science and Technology at Beijing Jiaotong University (China), with a research focus on Computer Vision. 
\end{IEEEbiography}

\vspace{-2em}

\begin{IEEEbiography}[{\includegraphics[width=1in,height=1.25in,clip,keepaspectratio]{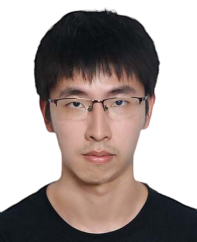}}]{Hongyu Pan}  received the B.E. degree from Beijing Institute of Technology (BIT) in 2016 and the M.S. degree in computer science from the Institute of Computing Technology (ICT), University of Chinese Academy of Sciences (UCAS), in 2019. He is currently an employee at Horizon Robotics. His research interests include computer vision, pattern recognition, and image processing. He specifically focuses on 3D detection/segmentation/motion and depth estimation.
\end{IEEEbiography}
\vspace{-2em}
\begin{IEEEbiography}
[{\includegraphics[width=1in,height=1.25in,clip,keepaspectratio]{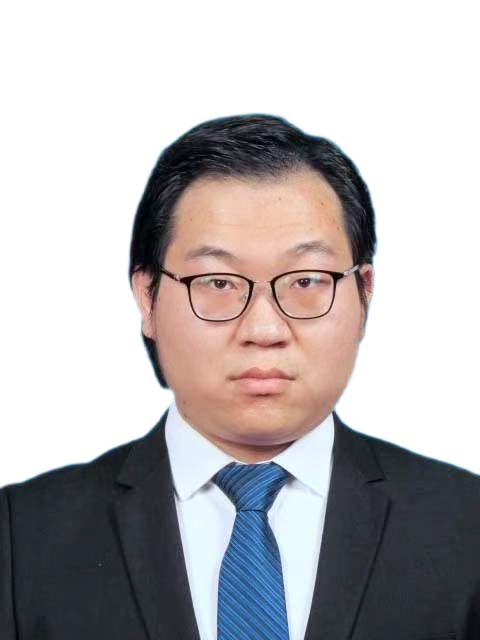}}]
{Feiyang Jia} was born in Yinchuan, Ningxia Province, China, in 1998. He received his B.S. degree from Beijing Jiaotong University (China) in 2020. He received a master's degree from Beijing Technology and Business University (China) in 2023. He is now a Ph.D. student majoring in Computer Science and Technology at Beijing Jiaotong University (China), with research focus on Computer Vision.
\end{IEEEbiography}

\vspace{-2em}

\begin{IEEEbiography}[{\includegraphics[width=1in,height=1.25in,clip,keepaspectratio]{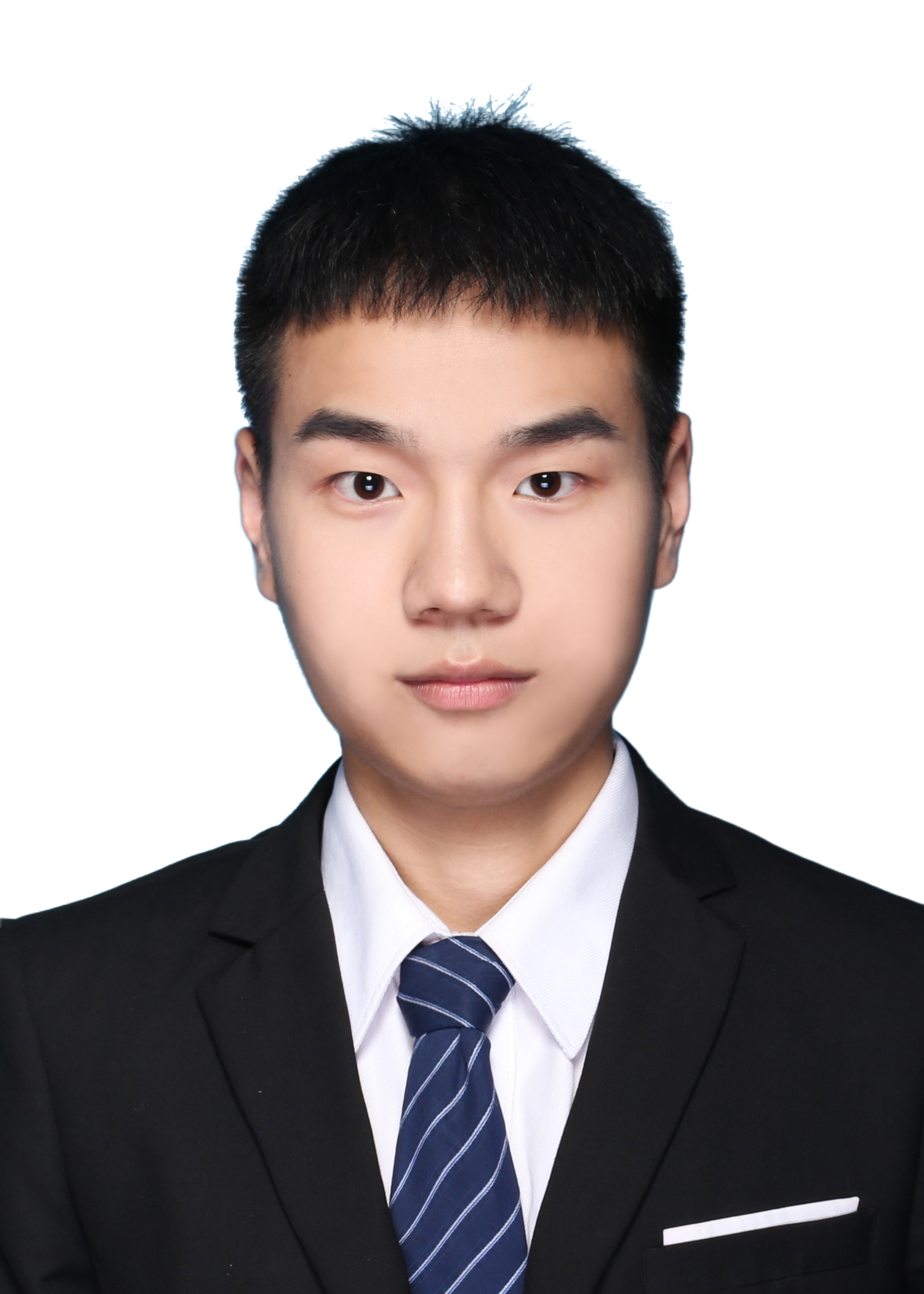}}]{Yongchang Zhang} 
received the B.E. degree from the University of Electronic Science and Technology of China (UESTC) in 2020 and the M.S. degree in control theory and control engineering from the Institute of Automation (CASIA), University of Chinese Academy of Sciences (UCAS) in 2023. He is currently a Research Engineer at Horizon Robotics. His research interests include computer vision, autonomous driving perception, and visual-language models. 
\end{IEEEbiography}
\vspace{-2em}

\begin{IEEEbiography}
[{\includegraphics[width=1in,height=1.25in,clip,keepaspectratio]{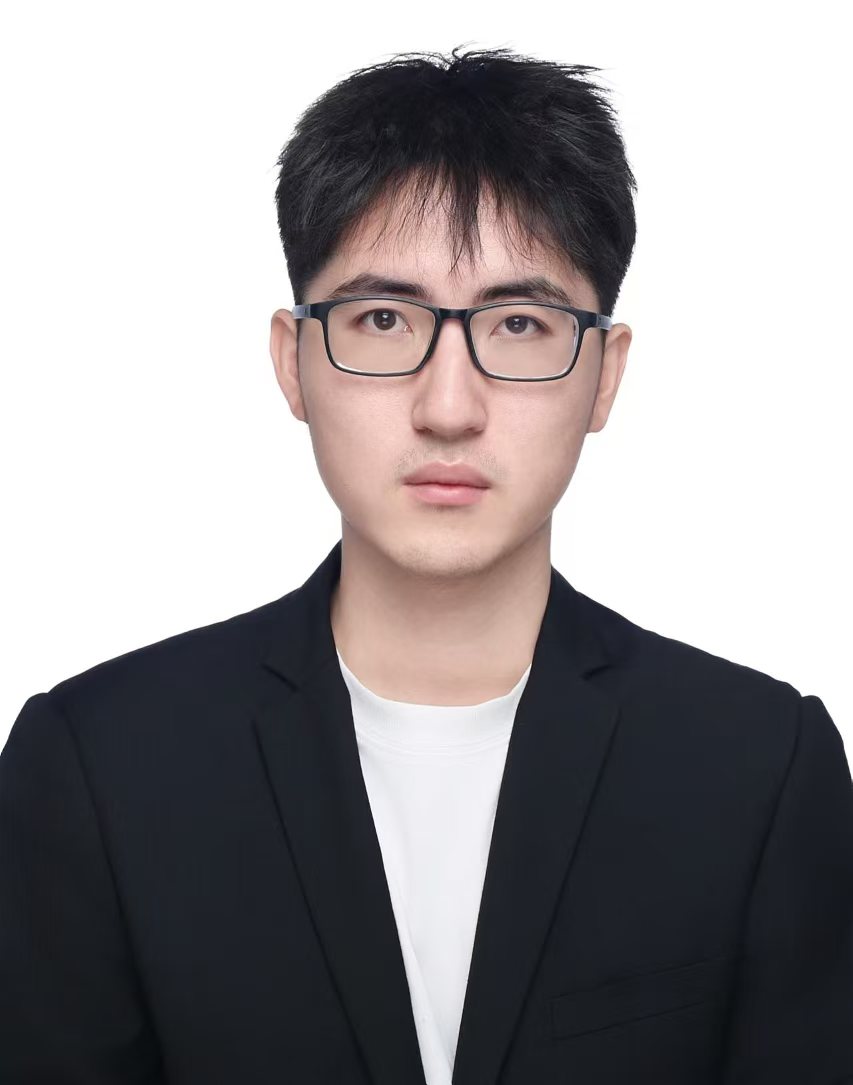}}] {Lin Liu} was born in Jinzhou, Liaoning Province, China, in 2001. He is now a college student majoring in Computer Science and Technology at China University of Geosciences(Beijing).
Since Dec. 2022, he has been recommended for a master's degree in Computer Science and Technology at Beijing Jiaotong University. His research interests are in computer vision.
\end{IEEEbiography}

\vspace{-2em}

\begin{IEEEbiography}[{\includegraphics[width=1in,height=1.25in,clip,keepaspectratio]{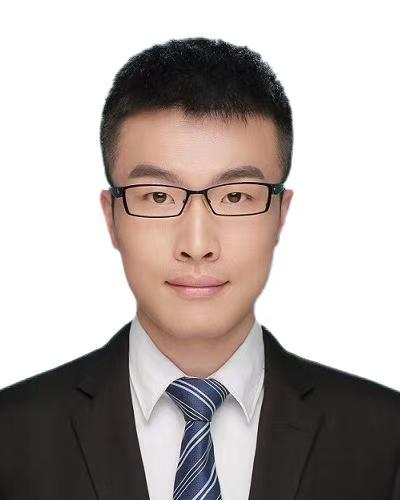}}]{Lei Yang} received his M.S. degree from the Robotics Institute at Beihang University in 2018. From 2018 to 2020, he worked as an algorithm researcher in JD.COM's Autonomous Driving R$\&$D Department. He obtained his Ph.D. degree from Tsinghua University's School of Vehicle and Mobility in 2024 and subsequently joined Nanyang Technological University, Singapore, as a Research Fellow in the School of Mechanical and Aerospace Engineering starting in 2025. His current research focuses on autonomous driving, 3D scene understanding, multi-modal large language model, and world model.
\end{IEEEbiography}
\vspace{-2em}

\begin{IEEEbiography}[{\includegraphics[width=1in,height=1.25in,clip,keepaspectratio]{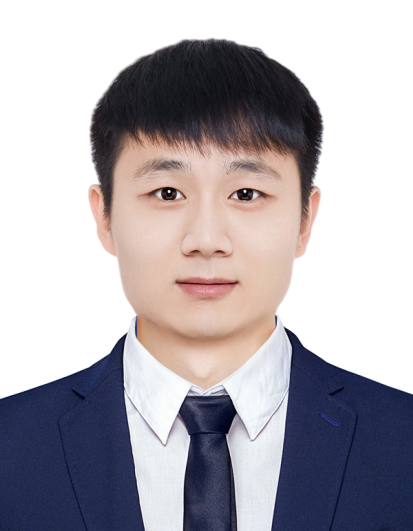}}]{Shaoqing Xu} received his M.S. degree in transportation engineering from the School of Transportation Science and Engineering in Beihang University. He is currently working toward the Ph.D. degree in electromechanical engineering with the State Key Laboratory of Internet of Things for Smart City, University of Macau, Macao SAR, China. His research interests include 3D Space Intelligence, End2End, WorldModel, VLA, and its applications in Autonomous Driving and Robotics.
\end{IEEEbiography}

\vspace{-2em}

\begin{IEEEbiography}[{\includegraphics[width=1in,height=1.25in,clip,keepaspectratio]{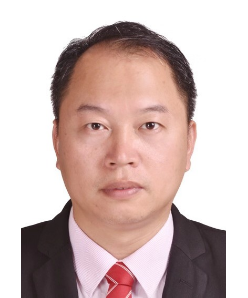}}]{Peiliang Wu} received the B.Sc. and Ph.D. degrees from Yanshan University, Qinhuangdao, China, in 2004 and 2010, respectively. He is currently a Postdoctoral Fellow with the the State Key Laboratory of Complex System Management and Control, Institute of Automation, Chinese Academy of Sciences, Beijing, China. He is also a Professor and Doctor Advisor with Yanshan University. He is a Member of the Academic Committee of Yanshan University, Member of the Standing Committee of the Youth Work Committee of the Chinese Artificial Intelligence Society, and the Vice Chairman of ACM Qinhuangdao. His research interests include robot learning and multi-agent systems.
\end{IEEEbiography}
\vspace{-2em}

\begin{IEEEbiography}[{\includegraphics[width=1in,height=1.25in,clip,keepaspectratio]{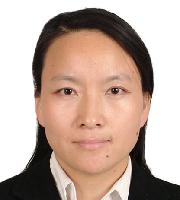}}]{Caiyan Jia}, born on March 2, 1976, is a lecturer and a postdoctoral fellow of the Chinese Computer Society. she graduated from Ningxia University in 1998 with a bachelor's degree in mathematics, Xiangtan University in 2001 with a master's degree in computational mathematics, specializing in intelligent information processing, and the Institute of Computing Technology of the Chinese Academy of Sciences in 2004 with a doctorate degree in engineering, specializing in data mining. she has received her D. degree in 2004. She is now a
professor in School of Computer Science and
Technology, Beijing Jiaotong University, Beijing,
China.
\end{IEEEbiography}
\vspace{-2em}

\begin{IEEEbiography}[{\includegraphics[width=1in,height=1.25in,clip,keepaspectratio]{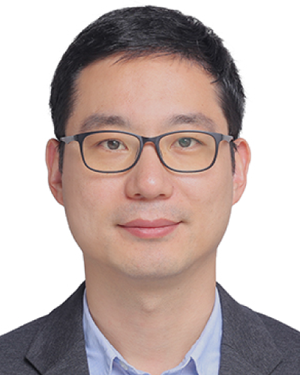}}]{Zheng Zhang}  (Senior Member, IEEE) received the Ph.D. degree from Harbin Institute of Technology, Shenzhen, China. He was a Post-Doctoral Research Fellow at The University of Queensland, St. Lucia, QLD, Australia. He is currently with Harbin Institute of Technology. He has authored or co-authored over 100 papers in leading international journals and conferences and received some prestigious recognitions. His main research interests include machine learning and multimedia. Dr. Zhang has regularly contributed as the Area Chair of ICML, NeurIPS, ICLR, CVPR, and ACM MM. He serves as an Associate Editor for IEEE TRANSACTIONS ON INFORMATION FORENSICS AND SECURITY, IEEE TRANSACTIONS ON AFFECTIVE COMPUTING, and IEEE JOURNAL OF BIOMEDICAL AND HEALTH INFORMATICS
\end{IEEEbiography}
\vspace{-2em}

\begin{IEEEbiography}[{\includegraphics[width=1in,height=1.25in,clip,keepaspectratio]{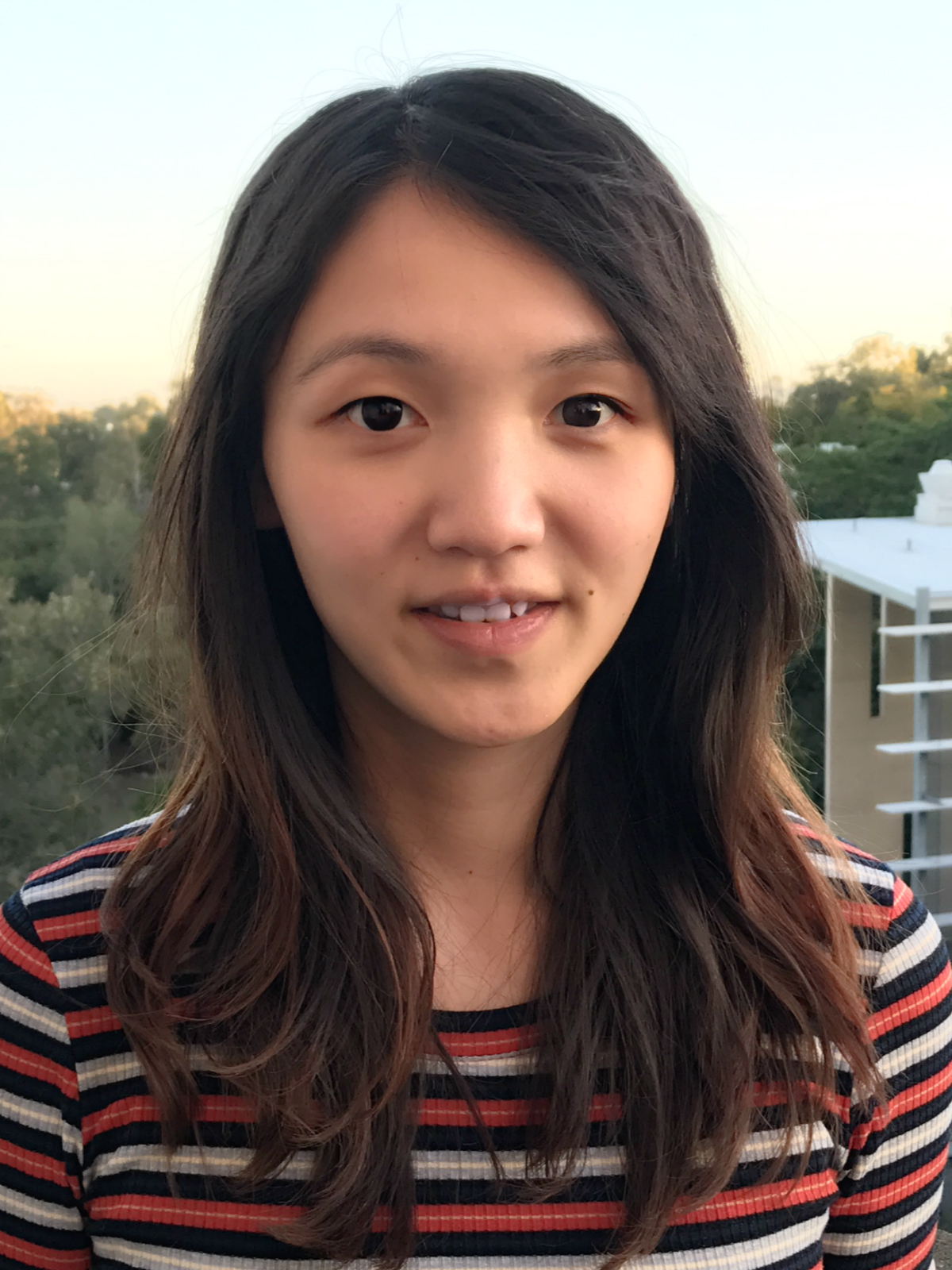}}]{Yadan Luo} (Member, IEEE) received the BS degree in computer science from the University of Electronic Engineering and Technology of China, and the PhD degree from the University of Queensland. Her research interests include machine learning, computer vision, and multimedia data analysis. She is now a lecturer with the University of Queensland.
\end{IEEEbiography}

\end{document}